\documentclass[12pt]{amsart}

\usepackage{natbib}
\usepackage{amscd,amssymb,amsmath,mathrsfs,amsfonts,graphicx,verbatim,epsfig,
psfrag, float, fancybox} % Typical maths resource packages
\usepackage{graphics}                 % Packages to allow inclusion of graphics
\usepackage{color}                    % For creating coloured text and background
\usepackage{hyperref}                 % For creating hyperlinks in cross references
\usepackage[TS1,OT1,T1]{fontenc}

\usepackage{amsmath}    % need for subequations
\usepackage{amscd,amssymb,mathrsfs,amsfonts}
\usepackage{graphicx}   % need for figures
\usepackage{verbatim}   % useful for program listings
\usepackage{color}      % use if color is used in tex\documentclass[10pt]{article}
\usepackage{amsmath}
\usepackage{verbatim}
\usepackage{setspace} 
\usepackage{algorithm}
\usepackage{easybmat}

\newcommand{\Keywords}[1]{\par\noindent
{\small{\em Keywords\/}: #1}}

\newcommand{\lpda}{$\ell_1$-PDA}

\begin{document}

\title{Discriminant analysis with adaptively pooled covariance}

\author{Noah Simon}
\author{Rob Tibshirani}

\begin{abstract}
Linear and Quadratic Discriminant analysis (LDA/QDA) are common tools for classification
problems. For these methods we assume
observations are normally distributed within group. We estimate a mean and covariance
matrix for each group and classify using Bayes theorem. With LDA, we estimate
a single, pooled covariance matrix, while for QDA we estimate a
separate covariance matrix for each group. Rarely do we believe in a homogeneous
covariance structure between groups, but often there is insufficient
data to separately estimate covariance matrices. We propose \lpda, a
regularized model which adaptively pools elements of the precision
matrices. Adaptively pooling these matrices decreases the
variance of our estimates (as in LDA), without overly biasing them. In this paper, we propose and discuss this method,
give an efficient algorithm to fit it for moderate sized problems, and show its efficacy on real and
simulated datasets. 
\\
\end{abstract}

\maketitle

\section{Introduction}\label{sec:intro}
Consider the usual two class problem: our data consists of $n$
observations, each observation with a known class label $\in
\{1,2\}$, and $p$ covariates measured per observation. Let $y$ denote
the $n$-vector corresponding to class (with $n_1$ observations in
class $1$ and $n_2$ in class $2$),
and $X$, the $n$ by $p$ matrix of covariates. We would
like to use this information to classify future observations.

We further assume that, given class $y(l)$, each observation, $x_l$, is independently normally distributed with some class specific mean
$\mu_{y(l)}\in\mathbb{R}^p$ and covariance $\Sigma_{y(l)}$, and that
$y(l)$ has prior probability $\pi_1$ of coming from class $1$ and
$\pi_2$ from class $2$. From here we estimate the two mean vectors,
covariance matrices, and prior probabilites and use these estimates
with Bayes theorem to classify future
observations. In the past a number of different methods have been
proposed to estimate these parameters. The simplest is Quadratic Discriminant
Analysis (QDA) which estimates the parameters by their maximum
likelihood estimates
\[
\pi_k = \frac{n_k}{n}
\] 
\[
\hat{\mu}_k = \frac{1}{n_k}\sum_{y(l) = k}x_l
\]
and
\[
\hat{\Sigma}_k = \frac{1}{n_k}\sum_{y(l) = k}\left(x_l - \mu_k\right)\left(x_l - \mu_k\right)^{\top}.
\]
To classify a new observation $x$, one finds the class with the highest
posterior probability. This is equivalent in the two class case to
considering
\begin{align*}
D(x) &= \operatorname{log}\left(\frac{\pi_1}{\pi_2}\right) -
\frac{1}{2}\left(x-\mu_1\right)^{\top}\Sigma_1^{-1}\left(x-\mu_1\right)\\
&+
\frac{1}{2}\left(x-\mu_2\right)^{\top}\Sigma_2^{-1}\left(x-\mu_2\right)
+ \operatorname{logdet}\left(\Sigma_1^{-1/2}\Sigma_2^{1/2}\right)
\end{align*}
and if $D(x)>0$ then classifying to class $2$, otherwise to class
$1$.\\

Linear Discriminant Analysis (LDA) is a similar but more commonly used
method. It estimates the parameters by a restricted MLE --- the
covariance matrices in both classes are constrained to be equal. So,
for LDA
\[
\hat{\Sigma}_1 = \hat{\Sigma}_2 = \frac{1}{n}\sum_{l=1}^n\left(x_l - \mu_{y(l)}\right)\left(x_l - \mu_{y(l)}\right)^{\top}
\]
While one rarely believes that the covariance matrices are exactly
equal, often the decreased variance from pooling the estimates greatly
outweights the increased bias.\\

\citet{friedman1989} proposed Regularized Discriminant Analysis (RDA)
noting that one can partially pool the covariance matrices and find a
more optimal bias/variance tradeoff.  He estimates $\Sigma_k$ by a
convex combination of the LDA and QDA estimates
\[
\hat{\Sigma}_k = \lambda\hat{\Sigma}_k^{\textrm{QDA}} + (1-\lambda)\hat{\Sigma}^{\textrm{LDA}}
\]
where $\lambda$ is generally determined by cross-validation.\\

We extend the idea of partially pooling the covariance matrices in a
different direction. We make the further
assumption that for most $i,j$, $\left(\Sigma_1^{-1}\right)_{i,j} \approx
\left(\Sigma_2^{-1}\right)_{i,j}$; that the partial covariance matrices are
mostly element-wise equal (or nearly equal). Intuitively this says that
conditional on all other variables, most pairs of
covariates interact identically in both groups.

Given this assumption, the natural approach is to find a restricted
MLE where the number of non-zero entries in $\Sigma_1^{-1} -
\Sigma_2^{-1}$ is constrained to be less than some $c$. ie. to find
\begin{align*}
\operatorname{argmax}\quad&\ell_1(\mu_1, \Sigma_1) + \ell_2(\mu_2,
\Sigma_2)\\
\textrm{s.t.} \quad& \left\|\Sigma_1^{-1} - \Sigma_2^{-1}\right\|_{0} \leq
c\\
& \Sigma_1, \Sigma_2 \textrm{ Positive Semi-Definite}
\end{align*}
where $\ell_k$ is the Gaussian log likelihood of the observations in
class $k$,
\[
\ell_k(\mu_k, \Sigma_k) = -\frac{n_k}{2}\operatorname{log}(2\pi) +
\frac{n_k}{2}\operatorname{logdet}\left(\Sigma_k^{-1}\right) +
\sum_{y(l) = k}\left(x_l - \mu_k\right)^{\top}\Sigma_k^{-1}\left(x_l - \mu_k\right)
\]
and $\|\cdot\|_{0}$ is the number of nonzero elements.
Unfortunately, this problem is not convex and would require a
combinatorial search. Instead we consider a convex relaxation
\begin{align}\label{eq:bound}
\operatorname{argmax}\quad&\ell_1(\mu_1, \Sigma_1) + \ell_2(\mu_2,
\Sigma_2)\\
\textrm{s.t.} \quad& \left\|\Sigma_1^{-1} - \Sigma_2^{-1}\right\|_{1} \leq
c\\
& \Sigma_1, \Sigma_2 \textrm{ Positive Semi-Definite}
\end{align}
where $\|\cdot\|_{1}$ is the sum of the absolute value of
the entries. Because the $|\cdot|$ is not differentiable at
$0$, solutions to \eqref{eq:bound} have few nonzero entries in
$\Sigma_1^{-1} - \Sigma_2^{-1}$ with the sparsity level dependent on
$c$. There is a large literature about using $\ell_1$
penalties to promote sparsity (\citet{tibs1996}, \citet{chen1996},
among others), and in particular sparsity has been applied in a
similar framework for graphical models \citep{BGA2008}. Also recently,
a very similar model to that which we propose has been applied to joint estimation of partial
dependence among many similar graphs \citep{danaher2011}. The astute
reader may note that \eqref{eq:bound} is not jointly convex in $\mu$ and $\Sigma^{-1}$.
However, we can still find the global maximum --- for fixed $\mu_1$ and $\mu_2$ it is convex, and, as we later
show, our estimates of $\mu_1$ and $\mu_2$ are completely independent
of our estimates of $\Sigma_1$, and $\Sigma_2$.\\

The problem~\eqref{eq:bound} has an equivalent Lagrangian form (which we
will write as a minimization for future convenience)
\begin{align}\label{eq:obj}
\operatorname{argmin}\quad& - \ell_1(\mu_1, \Sigma_1) - \ell_2(\mu_2,
\Sigma_2) + \lambda\left\|\Sigma_1^{-1} - \Sigma_2^{-1}\right\|_1\\
\textrm{s.t.}\quad & \Sigma_1, \Sigma_2 \textrm{ Positive Semi-Definite}
\end{align}
This is the objective which we will focus on in this
paper. We will call its solution ``$\ell_1$ Pooled Discriminant
Analysis'' (\lpda). For
$\lambda = 0$ these are just QDA estimates and for $\lambda$
sufficiently large, just LDA estimates.\\

In this paper, we examine the \lpda\, objective; we discuss the connections between \lpda\, and estimating
interactions in a logistic model; we show the efficacy of \lpda\,on real
and simulated data; and we give an efficient algorithm to fit \lpda\,based on the
alternating direction method of moments (ADMM).

\subsection{Reductions}
One may note that our objective~\eqref{eq:obj} is not jointly convex
in $\mu_k$ and $\Sigma_k$, however this is not a problem (the
optimization splits nicely). For a fixed $\Sigma_1$, $\mu_1^*$ minimizes
\[
\frac{1}{2}\sum_{y(l) =1}\left(x_l -
    \mu_1\right)^{\top}\Sigma_1^{-1}\left(x_l - \mu_1\right).
\]
This is true iff
\[
\Sigma_1^{-1}\sum_{y(l) =1}\left(x_l - \mu_1^*\right) = \underline{0}.
\]
Thus, $\mu_1^* = \bar{x}_1 = \frac{1}{n_1}\sum_{y(l) =1}x_l$ is the sample mean from
class $1$, and similarly $\mu_2^*$ is the sample mean from class
$2$. We can simplify our objective~\eqref{eq:obj} by substituting the
sample means in for $\mu_1^*$ and $\mu_2^*$ and noting that
\begin{align*}
\sum_{y(l) =1}\left(x_l -
    \bar{x}_1\right)^{\top}\Sigma_1^{-1}\left(x_l - \bar{x}_1\right) &
  = \frac{1}{2}\sum_{y(l) =1}\operatorname{tr}\left[\left(x_l -
    \bar{x}_1\right)^{\top}\Sigma_1^{-1}\left(x_l -
    \bar{x}_1\right)\right]\\
&= n_1\sum_{y(l) =1}\operatorname{tr}\left[\Sigma_1^{-1}\left(x_l -
    \bar{x}_1\right)\left(x_l -
    \bar{x}_1\right)^{\top}/n_1\right]\\
&=n_1\operatorname{tr}\left[\Sigma_1^{-1}\sum_{y(l) =1}\left(x_l -
    \bar{x}_1\right)\left(x_l -
    \bar{x}_1\right)^{\top}/n_1\right]\\
&=n_1\operatorname{tr}\left[\Sigma_1^{-1}S_1\right].
\end{align*}
where $\hat{\Sigma}_1$ is the sample covariance matrix for class
$1$.\\
Our new objective is
\begin{align}\label{eq:obj1}
\operatorname{min}_{\Sigma_1,\Sigma_2} & -n_1\operatorname{logdet}\left(\Sigma_1^{-1}\right) +
n_1\operatorname{tr}(\Sigma_1^{-1}S_1) -  n_2\operatorname{logdet}\left(\Sigma_2^{-1}\right)\\
& +n_2\operatorname{tr}(\Sigma_2^{-1}S_2) + \lambda ||\Sigma_1^{-1} -
  \Sigma_2^{-1}||_{1}
\end{align}
subject to $\Sigma_1$ and $\Sigma_2$ positive semi-definite
(PSD). This is a jointly convex problem in $\Sigma_1^{-1}$ and
$\Sigma_2^{-1}$.

\section{Solution Properties}
There is a vast literature on using $\ell_1$ norms to induce sparsity. In this section we will inspect the optimality conditions for our particular problem to gain some insight.  We begin by reparametrizing objective \eqref{eq:obj1} in terms of $\Delta = \left(\Sigma_1^{-1} - \Sigma_2^{-1}\right)/2$, and $\Theta = \left(\Sigma_1^{-1} + \Sigma_2^{-1}\right)/2$
\begin{align}\label{eq:obj1}
\operatorname{min}_{\Delta,\Theta} & -n_1\operatorname{logdet}\left(\Delta + \Theta\right) +
n_1\operatorname{tr}(\left[\Delta + \Theta\right]S_1) -  n_2\operatorname{logdet}\left(\Theta - \Delta\right)\\
& +n_2\operatorname{tr}(\left[\Theta - \Delta\right]S_2) + \lambda ||\Delta||_{1}
\end{align}
To find the Karush-Kuhn optimality conditions, we take the subgradient of this expression and set it equal to $0$. We see that
\begin{equation}\label{eq:opt1}
-n_1\left(\hat{\Delta} + \hat{\Theta}\right)^{-1} + n_1 S_1 -
n_2\left(\hat{\Theta} - \hat{\Delta}\right)^{-1} + n_2 S_2 + \lambda\partial(\hat{\Delta}) = 0 
\end{equation}
and
\begin{equation}\label{eq:opt2}
-n_1\left(\hat{\Delta} + \hat{\Theta}\right)^{-1} + n_1 S_1 + n_2\left(\hat{\Delta} - \hat{\Theta}\right)^{-1} - n_2 S_2 = 0 
\end{equation}
where $\hat{\Delta}$ and $\hat{\Theta}$ minimize the objective and $\partial(\Delta)$ is a $p$ by $p$ matrix with
\[
\partial(\Delta)_{i,j} =
\begin{cases}
\textrm{sign}(\Delta)_{i,j} , & \text{if } \Delta_{i,j} \neq 0\\
\in\left[-1,1\right], & \text{if } \Delta_{i,j} = 0
\end{cases}
\]
Now, we can substitute $\Sigma_1^{-1}$ and $\Sigma_2^{-1}$ back in to the subgradient equations:
\begin{equation}\label{eq:kkt1}
n_1\left(S_1 - \hat{\Sigma}_1\right) - n_2\left(S_2 - \hat{\Sigma}_2\right) + \lambda\partial(\hat{\Sigma}_1^{-1} - \hat{\Sigma}_2^{-1}) = 0 
\end{equation}
and
\begin{equation}\label{eq:kkt2}
S_{\textrm{pool}} \equiv \frac{n_1 S_1 + n_2 S_2}{n_1 + n_2} = \frac{n_1 \hat{\Sigma}_1 + n_2 \hat{\Sigma}_2}{n_1 + n_2}.
\end{equation}
 We find these optimality conditions curious as they directly involve $\hat{\Sigma}_k$ rather than $\hat{\Sigma}_k^{-1}$. Equation~\eqref{eq:kkt1} shows that the solution will have a sparse difference $\hat{\Sigma}_1^{-1} - \hat{\Sigma}_2^{-1}$. Though somewhat unintuitive, it parallels the KKT conditions for the Lasso and other $\ell_1$ penalized problems. In particular, because the subgradient of $\|\Delta\|_1$ can take a variety of values for $\Delta_{i,j} = 0$, the optimality conditions are often satisfied with $\Delta_{i,j} = 0$ for many $i,j$.
Equation~\eqref{eq:kkt2} shows us that the pooled average of our estimates is unchanged ($S_{\textrm{pool}} = \hat{\Sigma}_{\textrm{pool}}$). Given the form of our penalty we find it interesting that the pooled average of the $\hat{\Sigma}_k$ is constant (independent of $\lambda$) rather than some convex combination of the $\hat{\Sigma}_k^{-1}$.\\

From these optimality conditions one can easily find the optimal solutions at both ends of our path (for $\lambda=0$ and $\lambda$ sufficiently large). If $S_1$ and $S_2$ are full rank, then for $\lambda = 0$ the optimality conditions are satisfied by the QDA solution with $\partial = 0$, and for $\lambda > \lambda_{textrm{max}} \equiv {n_1n_2\left\|S_1 - S_2\right\|_{\infty}}/{(n_1 + n_2)}$ the conditions are satisfied by the LDA solution with $\partial = {n_1n_2(S_1 - S_2)}/{[\lambda(n_1+n_2)]}$. In Section~\ref{sec:opt}, we give a pathwise algorithm to fit \lpda\, along our path of $\lambda$-values from $\lambda_{\textrm{max}}$ to $0$.

\subsection{When is the problem ill posed?}
Recall that if $S_1$ or $S_2$ is not full rank, then the QDA solution is undefined. In our case one can see that as $\lambda\rightarrow 0$ we still have this difficulty, however for $\lambda > 0$, so long as $S_{\textrm{pool}} = \left(n_1S_1 + n_2S_2\right)/\left(n_1 + n_2\right)$ is full rank, our solution is well defined. In the case that $S_{\textrm{pool}}$ is not full rank, then the solution is ill-defined for all $\lambda$.

\section{Forward Vs Backward Model}
So far we have assumed a model in which the $x$-values are generated
given the class assignments. We will henceforth refer to this as the
``backward generative model'' or backward model. Many other approaches to
classification use a ``forward generative model'' wherein we consider
the class assignments to be generated from the x-values (eg. logistic
regression). Our backward model has a corresponding forward model. By
Bayes theorem we have
\begin{align*}
\operatorname{P}(y = 1 | x) &= \frac{\pi_1 \operatorname{exp}\left(l_1\right)}{\pi_2  \operatorname{exp}\left(l_2\right) + \pi_1
   \operatorname{exp}\left(l_1\right)}\\
&= \frac{ \operatorname{exp}\left[\operatorname{log}(\pi_1/\pi_2) +
    l_1 - l_2\right]}{1 +  \operatorname{exp}\left[\operatorname{log}(\pi_1/\pi_2) +
    l_1 - l_2\right]}
\end{align*}
where
\[
l_k = -(x-\hat{\mu}_k)^{\top}\hat{\Sigma}_k^{-1}(x-\hat{\mu}_k)/2.
\]
We can simplify this to get a better handle on it. Some algebra gives us
\begin{align}\label{eq:forward}
\operatorname{logit}\left[\operatorname{P}(y = 1 | x)\right] &= \operatorname{log}(\pi_1/\pi_2) + \mu_2^{\top}\Sigma_2^{-1}\mu_2/2 -
\mu_1^{\top}\Sigma_1^{-1}\mu_1/2\\ 
&+ \left(\Sigma_1^{-1}\mu_1 -
  \Sigma_2^{-1}\mu_2\right)^{\top} x + x^{\top}\left(\Sigma_2^{-1} -
    \Sigma_1^{-1}\right) x/2.
\end{align}
where $\operatorname{logit}(p) = p/(1-p)$. This is just a logistic model with interactions and
quadratic terms. In general a logistic model takes the form
\begin{equation*}
\operatorname{logit}\left[\operatorname{P}(y = 1 | x)\right] = \beta_0 + \sum \beta_i x_i + \sum_{i\leq j}\gamma_{i,j}
x_{i} x_{j}
\end{equation*}
or in matrix form
\begin{equation}\label{eq:logit}
\operatorname{logit}\left[\operatorname{P}(y = 1 | x)\right] =
\beta_0 + \beta^{\top} x + x^{\top}\Gamma x/2
\end{equation}
So our forward generative model in \eqref{eq:forward} is a logistic
model with
\begin{align*}
\beta_0 &= \operatorname{log}(\pi_1/\pi_2) + \mu_2^{\top}\Sigma_2^{-1}\mu_2/2 -
\mu_1^{\top}\Sigma_1^{-1}\mu_1/2\\
\beta &= \Sigma_1^{-1}\mu_1 -
  \Sigma_2^{-1}\mu_2\\
\Gamma/2 &= \Sigma_2^{-1} -
    \Sigma_1^{-1}
\end{align*}
Note, that with LDA we estimate $\Gamma$ to be identically $0$, with
QDA $\Gamma$ is entirely nonzero, and with \lpda , $\Gamma$ has both
zero and nonzero elements.

\subsection{Estimating Interactions}\label{sec:interactions}
Based on the forward model above, one can consider our sparse estimation of $\Gamma$ as a method for estimating sparse interactions. There has been a recent push to estimate interactions in the high
dimensional setting (\citet{radchenko2010}, \citet{zhao2009},
among others). The basic idea is to consider a general logistic model as
in \eqref{eq:logit} (or a linear model for continuous response), and
to estimate $\beta_0$, $\beta$, and $\Gamma$ in such a way that there are few nonzero entries in $\hat{\Gamma}$ (often the diagonal is constrained to be $0$). The simplest of these
approaches maximize a penalized logistic log-likelihood
\begin{align*}
\operatorname{argmax}_{\beta, \Gamma}\quad&\sum_{i=1}^n \big\{y(l) \operatorname{log}(p_l) +
(1-y(l))\operatorname{log}(1-p_l)\big\} -
\lambda||\Gamma||_1\\
\textrm{s.t.} \quad & \operatorname{log}\left(\frac{p_l}{1-p_l}\right) = \beta_0 + \beta^{\top} x_l + x_l^{\top}\Gamma x_l/2
\end{align*}
As we have shown, for discriminant analysis considered as a forward model, nonzero off-diagonal terms
in $\Gamma = \hat{\Sigma}_2^{-1} - \hat{\Sigma}_1^{-1}$ correspond to
pairs of variables with interactions. Thus \lpda\, estimates a logistic
model with sparse interactions (and quadratic terms). \lpda\, differs from other methods because it has additional distributional assumptions on the covariates which in turn put constraints on our estimates of $\beta_0$, $\beta$, and
$\Gamma$, but the underlying idea is the same.\\

\subsection{Linear Vs Quadratic Decision Boundaries}
The sparsity of $\Gamma$ again shows up if we consider the decision boundaries of discriminant analysis. For each method (LDA, QDA and \lpda), once the parameters are estimated, $\mathbb{R}^p$ is
partitioned into two connected spaces --- one space where the estimated posterior probability
of an observation is higher for class $1$ and another
space where it is higher for class $2$. The decision boundary is
$D =\left\{x\, |\, \operatorname{P}(y = 1 | x) =
0.5\right\}$ which is equivalent to $\left\{x\, |\, \operatorname{logit}\left[\operatorname{P}(y = 1 | x)\right] =
0\right\}$. Referring back to our forward generative framework, \eqref{eq:forward}, we see that
\[
D  =\left.\Big\{x\, \right |\, \hat{\beta}_0 + \hat{\beta}^{\top} x + x^{\top}\hat{\Gamma} x = 0\Big\}
\]
The nonzero terms in $\hat{\Gamma} = \hat{\Sigma}_2^{-1} - \hat{\Sigma}_1^{-1}$
correspond to pairs of dimensions in which the decision boundary is quadratic
rather than linear. As expected, LDA has a linear decision boundary, and QDA has a quadratic decision boundary (with all cross terms included). \lpda\,is a hybrid of these --- it is quadratic in some terms and linear in others.

\section{Comparisons}
A number of other methods have been proposed for discriminant analysis using
sparsity and pooling. These methods are useful, but fill a different role than \lpda. We will compare 2 of these ideas to \lpda\, and discuss when each is appropriate.
\subsection{RDA}
Regularized Discriminant Analysis \citep{friedman1989} estimates the
within class covariance matrices as a convex combination of the LDA and QDA estimates. Like \lpda it gives a path from LDA to QDA. In contrast RDA is
basis equivariant (changing the basis on which you train will not
change the predictions), while \lpda\,is not. In RDA, one uses a
common idea in empirical bayes and stein estimation --- we often
overestimate the magnitude of extreme effects, in our case we overestimate the
extremity of largest and smallest eigenvalues of $\Sigma_1 -
\Sigma_2$, so RDA shrinks these values. On the other hand,\lpda\,is very
basis specific. In \lpda, as in all
sparse signal processing, we believe we have a good basis (in our
case, we believe that the differences are sparse in this basis) and
would like to leverage this in our estimation.

\subsection{Sparse LDA}
A number of methods have been proposed for ``sparse LDA.''
(\citet{dudoit2002}, \citet{bickel2004}, \citet{witten2011}, among
others). These methods either assume diagonal covariance matrices and
look for sparse mean differences, or assume $\Sigma_1 = \Sigma_2$ and
(either implicitly or explicitly) look for sparsity in
$\Sigma^{-1}\left(\mu_1 - \mu_2\right)$. This gives a linear decision rule which uses only few
of the variables. These methods are well suited to very high
dimensional problems (they require many fewer observations than LDA).\\

In contrast \lpda\,does not remove variables --- it only shrinks
decision boundaries from quadratic to linear. It is not well suited to
very high dimensional problems. In particular, the solution is
degenerate if $p > n_1 + n_2$, but it will generally perform better than sparse LDA for $p < n_1 + n_2$.\\

To draw another parallel to logistic regression (as in
Section~\ref{sec:interactions}), Sparse LDA is similar to sparse
estimation of main effects (with no interactions), while \lpda\,is
similar to sparse estimation of interactions (with all main effects included).

\section{Optimization}\label{sec:opt}
One of the main attractions of this criterion is that it is a convex problem and hence a global optimum can be found relatively quickly. In particular we have developed a method
which can solve this for up to several hundred variables (though the accuracy in poorly conditioned larger problems can be an issue).\\

First, for ease of notation we introduce new variables: let $A =
\Sigma_1 ^{-1}$, $B = \Sigma_2^{-1}$, $S_A = S_1$, and $S_B
=  S_2$. If we plug
in the sample means for $\mu_1$ and $\mu_2$, our
new criterion (negated for convenience) is now
\begin{align}\label{eq:obj1}
\operatorname{min}_{A,B} & -n_1\operatorname{logdet}A +
n_1\operatorname{tr}(AS_A) -  n_2\operatorname{logdet}B\\
& +n_2\operatorname{tr}(BS_B) + \lambda ||A -
  B||_{1}
\end{align}
subject to $A$, $B$ PSD, where $n_1$ is the number of observations in group $1$, $n_2$ is the
number of observations in group $2$. Recall that this is convex in $A$ and $B$.\\

One could solve this using interior point methods discussed in
\citet{BV2004}. Unfortunately, for semi-definite programs the
complexity of interior point algorithms scales like $p^6$, making this
approach impractical for $p$ larger than $15$ or $20$. Instead we
develop an approach based on the alternating direction method of moments (ADMM) which scales up to several hundred covariates.

\subsection{ADMM Algorithm}
ADMM is an older class of algorithms which has recently seen a
re-emergence largely thanks to \citet{boyd2010}. Our particular algorithm is a
adaptation of their ADMM algorithm for sparse inverse
covariance estimation.  The motivation for this algorithm is simple --- the combination of a
logdet term and a $||\cdot||_1$ term makes our optimization difficult, so we split the $2$
up and introduce an auxiliary variable $C\equiv A - B$ and a dual variable $\Gamma$. We leave the details of
developing this algorithm to the appendix (though they are straightforward). The exact algorithm is
\begin{enumerate}
\item Initialize $A_0$, $B_0$, $C_0$, and $\Gamma_0$ and choose a fixed $\rho>0$
\item Iterate until convergence
\begin{enumerate}
\item Update $A$ by
\[
A_{k+1} = U\tilde{A}U^{\top}
\]
where $\rho\left(C_k + B_k +\Gamma_k\right) - n_1S_A =
UDU^{\top}$ is its eigenvalue decomposition (with $D = diag(d_i)$),
and $\tilde{A}$ is diagonal with
\[
\tilde{A}_{ii} = \frac{d_i + \sqrt{d_i^2 + 4\rho n_1}}{2\rho}
\]

\item Update $B$ by 
\[
B_{k+1} =  V\tilde{B}V^{\top}
\]
where  $\rho\left(A_{k+1} - C_k - \Gamma_k\right) - n_2S_B =
VEV^{\top}$ is its eigenvalue decomposition (with $E =
diag(e_i)$) and $\tilde{B}$ is diagonal with
\[
\tilde{B}_{ii} = \frac{e_i + \sqrt{e_i^2 + 4\rho n_2}}{2\rho}
\]
\item Update $C$ by
\[
C_{k+1} = \operatorname{S}_{\lambda/\rho}\left(A_{k+1} - B_{k+1} - \Gamma_k\right)
\]
where $S_{\lambda}(\cdot)$ is the element-wise soft thresholding operator
\[
S_{\lambda}(Z)_{i,j} = \textrm{sign}\left(Z_{i,j}\right)\operatorname{max}\left(\left|Z_{i,j}\right| - \lambda,\, 0\right)
\]
\item update $\Gamma$ by
\[
\Gamma_{k+1} = \Gamma_k + \rho \left(C_{k+1} - A_{k+1} + B_{k+1}\right)
\]
\end{enumerate}
\end{enumerate}
Upon convergence, $A^*$ and $B^*$ are the variables of interest (the
rest may be discarded). The complexity of each step of this algorithm is dominated by the eigenvalue decompositions, each of which require $O(p^3)$
operations. 

\section{Path-wise Solution}
Often we do not know a-priori what value our regularization parameter should be and would like to fit the entire path from $\lambda_{max}$ (corresponding to the LDA solution) to $\lambda = 0$ (corresponding to the QDA solution). We define 
\[
\lambda_{\max} \equiv \frac{n_1n_2\left\|S_1 - S_2\right\|_{\infty}}{n_1 + n_2}
\]
It is easy to see that for $\lambda \geq \lambda_{max}$, $\hat{\Sigma}_1 = \hat{\Sigma}_2 = \frac{n_1S_1 + n_2S_2}{n_1 + n_2}$ (our LDA solution) satisfies \eqref{eq:opt1} and \eqref{eq:opt2}, and thus is our solution. One can also see that $\hat{\Gamma} = \frac{n_1n_2\left(S_1 + S_2\right)}{n_1 + n_2}$ is our optimal dual variable for $\lambda \geq \lambda_{max}$.\\

To solve along a path we start at $\lambda = \lambda_{max}$, and plug in our known solution. We then decrease $\lambda$ and solve the new problem, initializing our algorithm at the previous $\hat{\Sigma}_1$, $\hat{\Sigma}_2$, and $\hat{\Gamma}$. Because $\lambda$ changes only slightly (and thus our solution changes only slightly), this approach is very efficient as compared to solving from scratch at each $\lambda$. When $S_A$ and $S_B$ are full rank our QDA solution is well defined and it is possible to run our path all the way to $\lambda=0$. Due to convergence issues along the potentially poorly conditioned end of the path (which we discuss in the next section) we instead choose to set $\lambda_{min} = \epsilon\lambda_{max}$ and log-linearly interpolate between the two (in our implementation default $\epsilon$ value is $0.01$).

\subsection{Convergence Issues}\label{sec:conv}
While ADMM is a good algorithm for finding an near exact solution, it is not considered a great algorithm for an exact solution (though it does eventually converge to arbitrary tolerance, this may require an unwieldy number of iterations). In our application, solving to machine tolerance is unnecessary (the statistical uncertainties are much greater than this). However, in some cases (especially with $p\sim n_1 + n_2$), near the end of the path our solution converges extremely slowly. Unfortunately there is no simple fix for this (more accurate interior point algorithms don't scale beyond $15$ or $20$ variables). While not ideal, this does not overly concern us --- convergence is slow in the region where $\Sigma_1^{-1} - \Sigma_2^{-1}$ is not very sparse (a region where we expect \lpda\, to perform poorly anyways). We will see an example of this issue arise later in Section~\ref{sec:real}.\\

One should also note that convergence rates near the end of the path are highly dependent on our choice of $\rho$. This is characteristic of all ADMM problems. To date, finding a disciplined choice of $\rho$ for ADMM is still an open question. We use $\rho = 1$ as our default, as it appears to work reasonably well for a range of problems.

\section{Simulated and Real Data}
To show its efficacy, we applied \lpda\,to real and simulated data.  In both cases we compare
our method to linear, quadratic and regularized discriminant analysis and show
improvement over both in overall classification error and on ROC
plots. In all problems \lpda\, was fit with $30$ lambda values log-linearly interpolating $\lambda_{\textrm{max}}$ and $0.01*\lambda_{\textrm{max}}$. RDA was fit with $30$ equally spaced $\lambda$-values between $0$ and $1$.

\subsection{Simulated Data}
We simulated data from the two class gaussian model described in
Section~\ref{sec:intro} with $p=30$ covariates.  We set $\Sigma_1 =
I_{p\times p}$ and 
\[
\Sigma_2 = \left(
\begin{BMAT}(c){c:c}{c:c}
C & 0\\
0 & I_{(p-5) \times (p-5)}
\end{BMAT}
\right)
\]
where $C$ is $5\times 5$ matrix with constant value $c$ on
the off diagonal entries, and $1$ on the diagonal. We also set a small mean
difference between the groups: $\mu_1 = \underline{0}_{p}$
\[
\mu_2 = \left(
\begin{BMAT}(c){c}{c:c}
\Delta \\
\underline{0}_{(p-10)} 
\end{BMAT}
\right)
\]
where $\Delta$ is a $10$-vector of $1$s\\

Under this model $\Sigma_1 ^{-1} -
\Sigma_2^{-1}$ is nonzero only in the upper left $5\times 5$
submatrix.    We simulated using varying numbers of
observations $n_1= n_2 \in(33,\,40,\,60)$, and values of $c \in
(0.3,\,0.6,\,0.9)$.  We used $3$ data sets for each simulation --- one
to fit the initial model, one to choose the optimal value of $\lambda$
and our final set to get an unbiased estimate of misclassification error.\\

\begin{table}[!h]
\begin{center}
\begin{tabular}{|l|l|c|c|c|}
\hline
& & \multicolumn{3}{l}{\# Observations}\\
& & \multicolumn{3}{l}{per Group ($n_k$)}\\
& & & &\\
& & $33$  & $40$  & $60$ \\
\hline
& & & &\\
c = 0.3& \lpda & 0.82 & 0.85  & 0.88\\
& LDA & 0.82 & 0.85 & 0.88\\
& QDA & 0.58 & 0.65 & 0.74\\
& RDA & 0.81 & 0.85 & 0.88\\
& & & &\\
\hline
& & & &\\
c = 0.6& \lpda & 0.82 & 0.83  & 0.87\\
& LDA & 0.82 & 0.84 & 0.86\\
& QDA & 0.59 & 0.66 & 0.76\\
& RDA & 0.81 & 0.83 & 0.86\\
& & & &\\
\hline
& & & &\\
c = 0.9& \lpda & 0.84 & 0.88  & 0.92\\
& LDA & 0.80 & 0.83 & 0.85\\
& QDA & 0.65 & 0.76 & 0.86\\
& RDA & 0.81 & 0.84 & 0.88\\
& & & &\\
\hline
\end{tabular}
\vspace{2 mm}
\caption[Comparison:sim]{Average $\%$ of correct classifications over $100$ simulated datasets (standard errors for all entries are less than $0.006$)}
\end{center}
\label{tab:1}
\end{table}

  \begin{figure}[t!]
  \centerline{
    \includegraphics[width=2.85in]{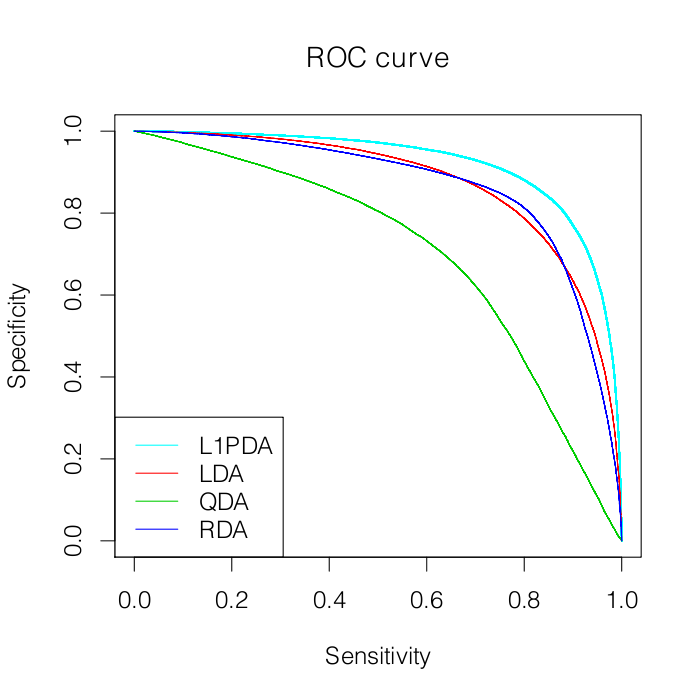}
  }
  \caption{Average ROC curve for simulated data with $n_k=33$, $c=0.9$}
  \label{fig:1}
  \end{figure}

As you can see from Table~\ref{tab:1}, when the signal to noise ratio (SNR)
is too small \lpda\,adaptively shrinks towards LDA and sees similar
performance.  When SNR is sufficiently large (the third group in the
table), \lpda\,is able to pick out the nonzero entries and achieves
substantially better misclassification rates. In these cases RDA also does fairly well (adaptively choosing
between LDA and QDA), however because it does not take sparsity into
account, it is outperformed by \lpda. We consider the large
SNR case more carefully in Figure~\ref{fig:1} (an ROC curve for $n_k=33$, $c =
0.9$) Again we used $3$ data sets per realization to get an unbiased curve estimate (and ran $100$ random realizations, though only average is shown on Figure~\ref{fig:1}). We estimated AUC for each procedure: \lpda\, $0.924\pm0.002$, LDA $0.875\pm 0.003$, QDA $0.732\pm 0.007$, and RDA $0.887\pm0.003$. \lpda\,does substantially better
than LDA, QDA, and RDA. With $p=30$ and $n_k=33$ there is clearly not enough
data for QDA to perform well (though the sample correlation matrices
are still invertible). However, as noted, \lpda\,also has a large edge
over LDA and RDA. \\

\section{Real Data}\label{sec:real}
We also applied \lpda\,to the ``Sonar, Mines vs. Rocks'' data
\citep{Gorman2011}.  This dataset has $60$ sonar signals measured on
each of $208$ objects (each labeled as either a rock or a mine).  We
randomly chose $150$ Mines/Rocks to train with, and then classified the remaining $58$.\\

  \begin{figure}[t!]
  \centerline{
    \mbox{\includegraphics[width=2.85in]{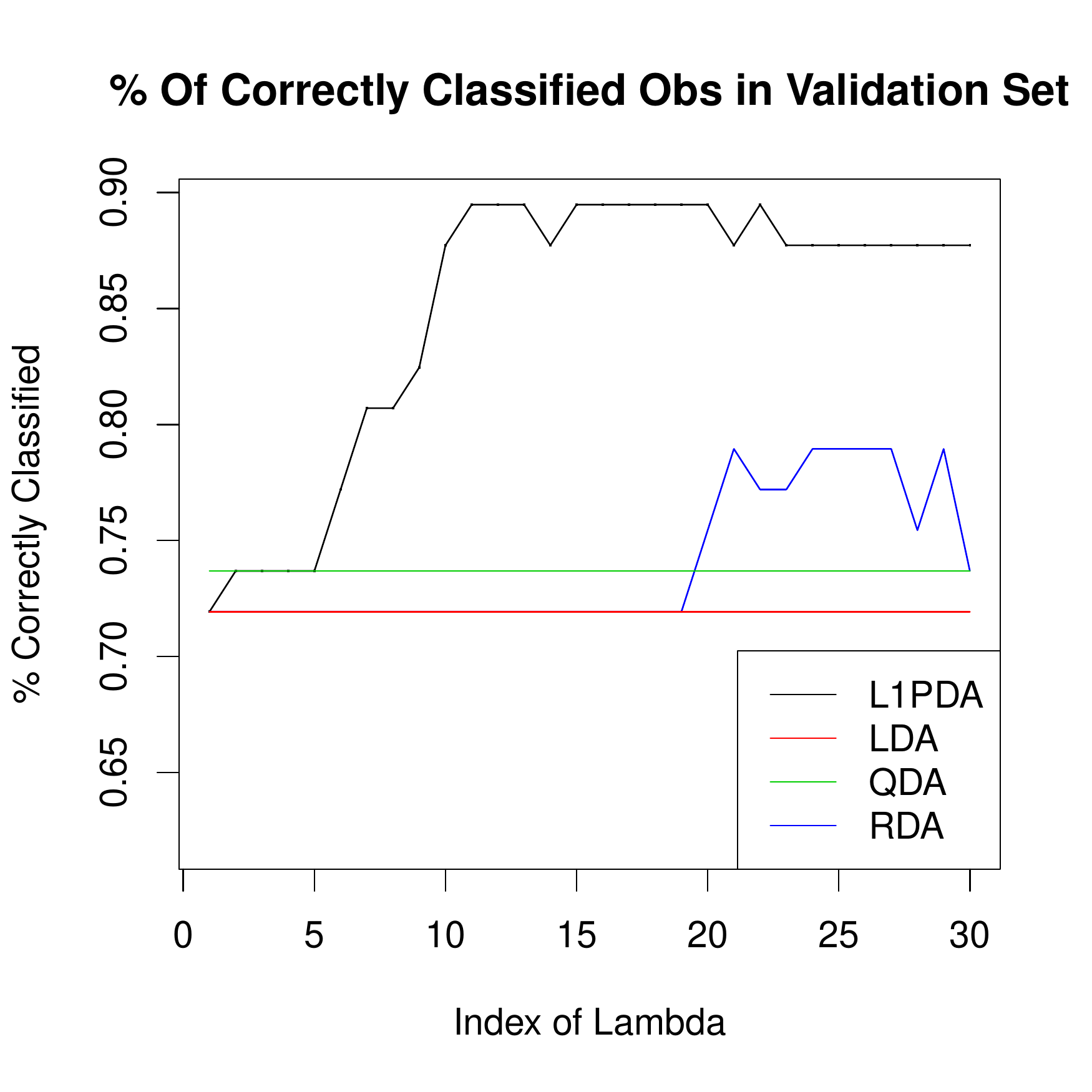}}
    \mbox{\includegraphics[width=2.85in]{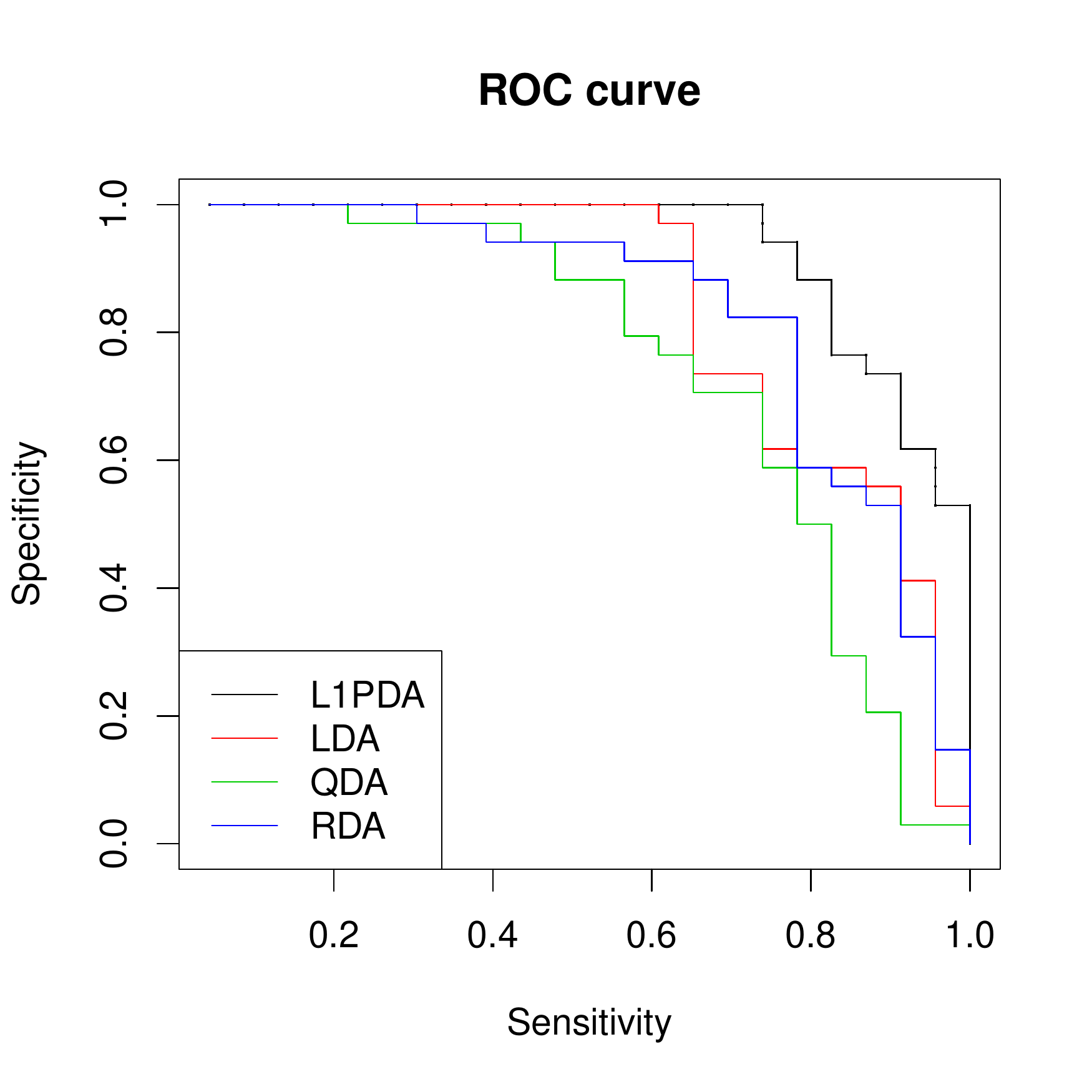}}
  }
  \caption{Plot of validated prediction accuracy for
    regularization path in $58$ mines/rocks, with $\lambda_{min} =
    0.01\lambda_{max}$ for \lpda, and ROC curve for $\lambda_{11}$}
  \label{fig:2}
  \end{figure}

As one can see from Figure~\ref{fig:2}, \lpda\,performs better on this
data than either LDA, QDA or RDA.  Estimated true classification rate peaks near the
middle of our regularization path, showing that a fair amount of
regularization can significantly improve classification. As we mentioned in Section~\ref{sec:conv} one can see convergence issues near the end of our path --- we would expect the CV error at our 30th $\lambda$-value to nearly match that of QDA (nearly rather than exactly because we don't run to $\lambda_{\textrm{min}} = 0$). However, it does not, indicating that our solution is not converging to the QDA solution. This does not overly concern us as our validation error reaches its crest well before this.\\
We also see
an ROC curve comparing \lpda, LDA, QDA, and RDA. For RDA we chose the simplest model which maximized predictive accuracy (the $21$st $\lambda$
value), and for \lpda\,the tenth $\lambda$ value, the most regularized
model before a precipitous drop in predictive accuracy (so as to
minimize bias for \lpda). The \lpda\,curve may still be slightly biased as we chose it from a section of our
path seen to do well in overall classification error (though not the
peak). Nonetheless, this curve appears indicative of an advantage
from \lpda\,over LDA, QDA, and RDA.

\section{Discussion}
In this paper we proposed \lpda, a classification method for
gaussian data which adaptively pools the precision matrices. We
motivated our method, and showed connections between it and estimating
sparse interactions. We gave two efficient algorithms to fit have
\lpda, and have shown its efficacy on real and simulated
data. We have made and plan to provide an {\tt R} implementation for \lpda\, publically available on CRAN. 

\section{Appendix $\alpha$}
We include a short overview of the ADMM algorithm. We can rewrite \eqref{eq:obj1} as 
\begin{align*}
\operatorname{min}_{A,B} & -n_1\operatorname{logdet}A +
n_1\operatorname{tr}(AS_A) -  n_2\operatorname{logdet}B\\
& +n_2\operatorname{tr}(BS_B) + \lambda ||C||_{1}\\
\textrm{s.t. }&\, C = A - B 
\end{align*}
 At the optimum we have $C = A - B$, so this is equivalent to
\begin{align}\label{eq:strongConv}
\operatorname{min}_{A,B,C}\, & -n_1\operatorname{logdet}A +
n_1\operatorname{tr}(AS_A) -  n_2\operatorname{logdet}B\\
& +n_2\operatorname{tr}(BS_B) + \lambda ||C||_{1} +
\frac{\rho}{2} \left\|C - A + B\right\|_F^2\\
\textrm{s.t. }&\, C = A - B 
\end{align}
$\rho$ can be any fixed positive number (though its choice will
affect the convergence rate of algorithm). We will motivate this
addition shortly. Now, using strong
duality, we can move our contraint into the objective, and finally
arrive at
\begin{align}\label{eq:admmObj}
\operatorname{max}_{\Gamma}\operatorname{min}_{A,B,C}\, &-n_1\operatorname{logdet}A +
n_1\operatorname{tr}(AS_A) -  n_2\operatorname{logdet}B\\
& +n_2\operatorname{tr}(BS_B) + \lambda ||C||_{1} +
\rho\operatorname{trace}\left[\Gamma^{\top}\left(C - A + B\right)\right]\\
&+ \frac{\rho}{2} \left\|C - A + B\right\|_F^2
\end{align}
For ease of notation we denote
\begin{align*}
\psi_{\Gamma}(A,B,C) &= -n_1\operatorname{logdet}A +
n_1\operatorname{tr}(AS_A) -  n_2\operatorname{logdet}B\\
& +n_2\operatorname{tr}(BS_B) + \lambda ||C||_{1} +
\rho\operatorname{trace}\left[\Gamma^{\top}\left( C - A + B\right)\right]\\
&+ \frac{\rho}{2} \left\|C - \left( A - B\right)\right\|_F^2
\end{align*}
and
\[
M(\Gamma) = \operatorname{min}_{A,B,C}\psi_{\Gamma}(A,B,C).
\]
Now, by basic convex analysis, the dual of any strongly convex function (with convexity constant
$\rho$) is differentiable and its derivative has lipschitz constant
$\rho$. Unfortunately \eqref{eq:strongConv} is not necessarily
strongly convex, however the addition of $||C-A+B||_F^2$, affords it
many of the same properties.
In particular if $C^*,\,A^*,\,B^*$ are the argmin of $\psi_{\Gamma_0}$ for a given
$\Gamma_0$, then 
\[
\left.\frac{\partial}{\partial \Gamma}M \right|_{\Gamma_0} = C^* - A^* + B^*
\]
If we could easily calculate $M(\Gamma)$, then we could use gradient
ascent on $\Gamma$
\[
\Gamma_{k+1} = \Gamma_k + \rho(C_k^* - A_k^* + B_k^*)
\]
 and one would have $A_k^*$ and $B_k^*$ converging to the argmax of our
original problem \eqref{eq:obj}. Unfortunately, $M(\Gamma)$ is not
easy to calculate, however $\psi_{\Gamma}$ is relatively simple to
minimize in one variable at a time ($A$, $B$, or $C$) with all
other variables fixed.  In ADMM we employ the same idea as gradient
descent, only we fudge the details --- instead of actually calculating
$M(\Gamma)$, we minimize first in $A$, with $B$, and $C$ fixed, then
in $B$ with $A$ and $C$ fixed and finally in $C$ with $A$ and $B$
fixed.  After these $3$ updates, we take our ``gradient'' step as
before (though this time it is not a true gradient step). This
leads to the following algorithm:
\begin{enumerate}
\item Initialize $A_0$, $B_0$, $C_0$, and $\Gamma_0$
\item Iterate until convergence
\begin{enumerate}
\item Update $A$ by
\[
A_{k+1} = \operatorname{argmin}_A \psi_{\Gamma_k}(A,B_k,C_k)
\]
\item Update $B$ by 
\[
B_{k+1} = \operatorname{argmin}_B \psi_{\Gamma_k}(A_{k+1},B,C_k)
\]
\item Update $C$ by
\[
C_{k+1} = \operatorname{argmin}_C \psi_{\Gamma_k}(A_{k+1},B_{k+1},C)
\]
\item Take ``gradient step''; update $\Gamma$ by
\[
\Gamma_{k+1} = \Gamma_k + \rho \left(C_{k+1} - A_{k+1} + B_{k+1}\right)
\]
\end{enumerate}
\end{enumerate}
One may note that if we instead iterate steps $a-c$ to convergence each time
before taking step $d$, we end up again with gradient descent.

\subsection{Inner Loop Updates}
In this section we derive the exact updates for $A$, $B$, and $C$ in
steps $a,\,b$ and $c$ of our ADMM algorithm. We begin with $A$: to
find $\operatorname{argmin}_A \psi_{\Gamma_k}(A,B_k,C_k)$ we must
minimize
\begin{align*}
&-n_1\operatorname{logdet}A +
n_1\operatorname{tr}(AS_A) +
\rho\operatorname{trace}\left[\Gamma_k^{\top}\left( C_k - A + B_k\right)\right]\\
&+ \frac{\rho}{2} \left|C_k - A + B_k\right\|_F^2
\end{align*}
If we take the derivative of this and set it equal to $0$ we get
\begin{equation}\label{eq:opt}
\rho A - n_1 A^{-1} = \rho\left(C_k + B_k +\Gamma_k\right) - n_1S_A
\end{equation}
Now if we let $\rho\left(C_k + B_k +\Gamma_k\right) - n_1S_A =
UDU^{\top}$ be its eigenvalue decomposition (with $D = diag(d_i)$),
then \eqref{eq:opt} is satisfied by
\[
A = U\tilde{A}U^{\top}
\]
where $\tilde{A}$ is diagonal and
\[
\tilde{A}_{ii} = \frac{d_i + \sqrt{d_i^2 + 4\rho n_1}}{2\rho}
\]
We can solve for $B_{k+1}$ similarly. Let $\rho\left(A_{k+1} - C_k - \Gamma_k\right) - n_2S_B =
VEV^{\top}$ be its eigenvalue decomposition (with $E =
diag(e_i)$). Then $\operatorname{argmin}_B
\psi_{\Gamma_k}(A_{k+1},B_k,C_k)$ is
\[
B = V\tilde{B}V^{\top}
\]
where $\tilde{B}$ is diagonal and
\[
\tilde{B}_{ii} = \frac{e_i + \sqrt{e_i^2 + 4\rho n_2}}{2\rho}
\]
The last variable to solve for is $C$.  Ignoring all terms without a
$C$, we need to minimize
\[
\lambda ||C||_{1} +
\rho\operatorname{trace}\left[\Gamma^{\top}\left( C - A +
    B\right)\right] + \frac{\rho}{2} \left\|C - A +
    B\right\|_F^2
\]
This is equivalent to minimizing
\[
\frac{1}{2}\left\|C - \left(A_{k+1} - B_{k+1} - \Gamma\right)\right\|_F^2 + \frac{\lambda}{\rho}||C||_{1}
\]
which is solved by
\[
C = \operatorname{S}_{\lambda/\rho}\left(A_{k+1} - B_{k+1} - \Gamma\right)
\]
where $\operatorname{S}_{\lambda/\rho}$ is the
entry-wise soft thresholding operator on the entries of the matrix. For $i\neq j$
\[
S_{\lambda/\rho}(X)_{ij} =
\operatorname{sign}\left(X_{ij}\right)\operatorname{max}\left(\left|X_{ij}\right|
  - \lambda/\rho, 0\right)
\]

So, in full detail, our algorithm is
\begin{enumerate}
\item Initialize $A_0$, $B_0$, $C_0$, and $\Gamma_0$
\item Iterate until convergence
\begin{enumerate}
\item Update $A$ by
\[
A_{k+1} = U\tilde{A}U^{\top}
\]
where $\rho\left(C_k - B_k +\Gamma_k\right) - n_1S_A =
UDU^{\top}$ is its eigenvalue decomposition (with $D = diag(d_i)$),
and $\tilde{A}$ is diagonal with
\[
\tilde{A}_{ii} = \frac{d_i + \sqrt{d_i^2 + 4\rho n_1}}{2\rho}
\]

\item Update $B$ by 
\[
B_{k+1} =  V\tilde{B}V^{\top}
\]
where  $\rho\left(A_{k+1} - C_k - \Gamma_k\right) - n_2S_B =
VEV^{\top}$ is its eigenvalue decomposition (with $E =
diag(e_i)$) and $\tilde{B}$ is diagonal with
\[
\tilde{B}_{ii} = \frac{e_i + \sqrt{e_i^2 + 4\rho n_2}}{2\rho}
\]
\item Update $C$ by
\[
C_{k+1} = \operatorname{S}_{\lambda/\rho}\left(A_{k+1} - B_{k+1} - \Gamma_k\right)
\]
\item update $\Gamma$ by
\[
\Gamma_{k+1} = \Gamma_k + \rho \left(C_{k+1} - A_{k+1} + B_{k+1}\right)
\]
\end{enumerate}
\end{enumerate}
The complexity of each step of this algorithm is dominated by the
eigenvalue decompositions, each of which require $O(p^3)$
operations. For this reason, while the algorithm can solve problems
for $p$ in the hundreds, it will be difficult to scale to larger
problems. One should note that $p$ in the hundreds is already an
optimization problem with tens of thousands of variables.

\bibliographystyle{abbrvnat}
\bibliography{/home/nsimon/texlib/simon}

\begin{thebibliography}{13}
\providecommand{\natexlab}[1]{#1}
\providecommand{\url}[1]{\texttt{#1}}
\expandafter\ifx\csname urlstyle\endcsname\relax
  \providecommand{\doi}[1]{doi: #1}\else
  \providecommand{\doi}{doi: \begingroup \urlstyle{rm}\Url}\fi

\bibitem[Banerjee et~al.(2008)Banerjee, Ghaoui, and d'Aspremont]{BGA2008}
O.~Banerjee, L.~E. Ghaoui, and A.~d'Aspremont.
\newblock Model selection through sparse maximum likelihood estimation for
  multivariate gaussian or binary data.
\newblock \emph{Journal of Machine Learning Research}, 9:\penalty0 485--516,
  2008.

\bibitem[Bickel and Levina(2004)]{bickel2004}
P.~Bickel and E.~Levina.
\newblock Some theory for fisher's linear discriminant function,'naive bayes',
  and some alternatives when there are many more variables than observations.
\newblock \emph{Bernoulli}, pages 989--1010, 2004.

\bibitem[Boyd and Vandenberghe(2004)]{BV2004}
S.~Boyd and L.~Vandenberghe.
\newblock \emph{Convex Optimization}.
\newblock Cambridge University Press, 2004.

\bibitem[Boyd et~al.(2010)Boyd, Parikh, Chu, Peleato, and Eckstein]{boyd2010}
S.~Boyd, N.~Parikh, E.~Chu, B.~Peleato, and J.~Eckstein.
\newblock Distributed optimization and statistical learning via the alternating
  direction method of multipliers.
\newblock \emph{Machine Learning}, 3\penalty0 (1):\penalty0 1--123, 2010.

\bibitem[Chen et~al.(1998)Chen, Donoho, and Saunders]{chen1996}
S.~S. Chen, D.~L. Donoho, and M.~A. Saunders.
\newblock Atomic decomposition by basis pursuit.
\newblock \emph{SIAM Journal on Scientific Computing}, pages 33--61, 1998.

\bibitem[Danaher et~al.(2011)Danaher, Wang, and Witten]{danaher2011}
P.~Danaher, P.~Wang, and D.~Witten.
\newblock The joint graphical lasso for inverse covariance estimation across
  multiple classes.
\newblock \emph{Arxiv preprint arXiv:1111.0324}, 2011.

\bibitem[Dudoit et~al.(2002)Dudoit, Fridlyand, and Speed]{dudoit2002}
S.~Dudoit, J.~Fridlyand, and T.~Speed.
\newblock Comparison of discrimination methods for the classification of tumors
  using gene expression data.
\newblock \emph{Journal of the American statistical association}, 97\penalty0
  (457):\penalty0 77--87, 2002.

\bibitem[Friedman(1989)]{friedman1989}
J.~Friedman.
\newblock Regularized discriminant analysis.
\newblock \emph{Journal of the American statistical association}, pages
  165--175, 1989.

\bibitem[Gorman and Sejnowski(2010)]{Gorman2011}
R.~Gorman and T.~Sejnowski.
\newblock Uci: Machine learning repository, 2010.
\newblock URL \url{http://archive.ics.uci.edu/ml}.

\bibitem[Radchenko and James(2010)]{radchenko2010}
P.~Radchenko and G.~James.
\newblock Variable selection using adaptive nonlinear interaction structures in
  high dimensions.
\newblock \emph{Journal of the American Statistical Association}, 105\penalty0
  (492):\penalty0 1541--1553, 2010.

\bibitem[Tibshirani(1996)]{tibs1996}
R.~Tibshirani.
\newblock Regression shrinkage and selection via the lasso.
\newblock \emph{Journal of the Royal Statistical Society B}, 58:\penalty0
  267--288, 1996.

\bibitem[Witten and Tibshirani(2011)]{witten2011}
D.~Witten and R.~Tibshirani.
\newblock Penalized classification using fisher’s linear discriminant.
\newblock \emph{Journal of the Royal Statistical Society, Series B}, 2011.

\bibitem[Zhao et~al.(2009)Zhao, Rocha, and Yu]{zhao2009}
P.~Zhao, G.~Rocha, and B.~Yu.
\newblock The composite absolute penalties family for grouped and hierarchical
  variable selection.
\newblock \emph{The Annals of Statistics}, 37\penalty0 (6A):\penalty0
  3468--3497, 2009.

\end{thebibliography}

\end{document}